\documentclass[runningheads]{llncs}

\usepackage{eccvabbrv}

\usepackage{graphicx}
\usepackage{booktabs}
\usepackage{amsmath}
\usepackage{amssymb}
\usepackage{makecell}
\usepackage{multirow}
\usepackage{xcolor}

\ifx\XeTeXversion\undefined
\usepackage[accsupp]{axessibility}  
\fi

\usepackage{orcidlink}

\newcommand{\metric}[4]{\ensuremath{#1_{\scriptstyle \pm #2}\,/\,#3_{\scriptstyle \pm #4}}}
\newcommand{\probe}[4]{\ensuremath{#1_{\scriptstyle \pm #2}\,/\,#3_{\scriptstyle \pm #4}}}

\newcommand{\dropstat}[3]{\ensuremath{#1_{\scriptstyle \pm #2}\,/\,#3}}
\newcommand{\gatepred}[4]{\ensuremath{#1_{\scriptstyle \pm #2}\,/\,#3_{\scriptstyle \pm #4}}}
\newcommand{\dirres}[4]{\ensuremath{#1_{\scriptstyle \pm #2}\,/\,#3_{\scriptstyle \pm #4}}}
\newcommand{\labelsel}[4]{\ensuremath{#1_{\scriptstyle \pm #2}\,/\,#3_{\scriptstyle \pm #4}}}

\newcommand{\restpair}[4]{\ensuremath{#1_{\scriptstyle \pm #2}\,/\,#3_{\scriptstyle \pm #4}}}
\newcommand{\restpairbold}[4]{\ensuremath{\mathbf{#1}_{\scriptstyle \pm #2}\,/\,#3_{\scriptstyle \pm #4}}}

\definecolor{badred}{RGB}{180,45,45}
\definecolor{goodgreen}{RGB}{30,140,70}

\begin{document}

\title{Association Restoration Test: Revealing Restorable Shortcuts after Unlearning } 

\titlerunning{Association Restoration Test}

\author{Amy Lu\thanks{These authors contributed equally to this work.} \and 
Changxiu Ji$^\star$}
\authorrunning{A. Lu and C. Ji}

\institute{
Department of Computer Science, Stanford University\\
\email{\{amylucky, leo0610\}@stanford.edu}
}

\maketitle

\begin{abstract}
Association unlearning aims to disable learned label--attribute shortcuts while preserving task performance. Existing evaluations mainly measure output-level robustness or probe whether
shortcut attributes remain readable in frozen features, but neither test
determines whether a retained association remains functionally usable by the
original classifier. We propose the Association Restoration Test (ART), a post-hoc diagnostic for
functional shortcut restorability. ART estimates class-conditional association directions, amplifies residual
components, and evaluates the modified features with the original classifier
head.
Across Waterbirds, CelebA, SpuCoDogs, and an ISIC timestamp-artifact extension, we show that output metrics, representation probes, and ART characterize distinct aspects of shortcut mitigation. These findings motivate restoration-aware evaluation for unlearning and
shortcut-mitigation methods that target learned associations rather than
individual classes or concepts.
\end{abstract}

\section{Introduction}
\label{sec:intro}

Machine unlearning~\cite{cao2015towards,bourtoule2021machine} typically removes
specified samples, classes, or concepts from a trained model while preserving
performance on the remaining task~\cite{fan2024salun}. Spurious-correlation
settings raise a different target: not a class or attribute itself, but a
learned relationship between the target label and an attribute. For example, Waterbirds correlates bird type with background, and CelebA-Blond
correlates hair color with gender~\cite{sagawa2020groupdro,koh2021wilds,liu2015celeba}.
We refer to this evaluation setting as \emph{association unlearning}: disabling
a learned label--attribute shortcut while preserving the task label.

Evaluating association unlearning requires separating output behavior,
representational readability, and functional use. Output metrics such as worst-group accuracy (WGA) measure whether the model
currently avoids shortcut-driven errors, but not whether the shortcut remains
encoded internally. Representation-level audits probe frozen features to test whether supposedly
removed information remains readable~\cite{gao2026illusion,yu2026mirage}. However, readability is not functional use: a feature can remain decodable
without being used by the original classifier head.

Restoration-aware evaluation has recently exposed similar failures in
class-unlearning settings, where models can satisfy output-level forgetting
metrics while retaining recoverable class information
~\cite{george2025illusion,gao2026illusion,qiu2025illusionforgetting}. These
works recover forgotten behavior using lightweight probes, relearning, or
feature steering~\cite{ha2025blindspots,qiu2025illusionforgetting,jang2026suppression,gao2026illusion}. However, class restoration tests recovery
of a forgotten class, whereas association restoration tests reactivation of a
learned label--attribute shortcut.

To fill this gap, we propose the \emph{Association Restoration Test}
(\textsc{ART}), a post-hoc feature-space diagnostic for functional
restorability of label--attribute associations. \textsc{ART} estimates class-conditional association directions in frozen
feature space, amplifies each example's residual component along the
corresponding direction, and applies the original classifier head to the modified feature to obtain the restored prediction. We use \textsc{ART} to audit visual shortcut-mitigation methods across
Waterbirds, CelebA, SpuCoDogs, and a multiclass ISIC timestamp-artifact
extension~\cite{sagawa2020groupdro,koh2021wilds,liu2015celeba,joshi2023spuco,tschandl2018ham10000,codella2018skin,hernandez2024bcn20000}.
We audit a spectrum of methods that reduce or suppress
label--attribute shortcuts, including group-robust debiasing references,
post-hoc classifier reweighting methods, and association-adapted unlearning
mechanisms.

Our audit shows that output robustness, feature readability, and functional
restorability can diverge. Shortcut attributes often remain readable from frozen
features even when WGA improves. ART further separates readable-but-decoupled
shortcuts from restorable shortcuts: some methods appear to decouple the
classifier head from retained shortcut structure, whereas others leave
associations that ART can reactivate, causing WGA drops and increased
shortcut-consistent errors.

Overall, our contributions are:
\begin{itemize}
\item \textbf{Association Restoration Test.}
We introduce \textsc{ART}, a post-hoc diagnostic for testing whether retained
label--attribute associations remain functionally restorable under the original
classifier head.

\item \textbf{Association-level restoration auditing.}
We extend restoration-aware evaluation from class-level unlearning to learned
label--attribute shortcuts.

\item \textbf{Empirical evidence of evaluation divergence.}
Across visual shortcut benchmarks, we show that output robustness, feature
readability, and functional restorability can diverge across shortcut-mitigation
and association-adapted unlearning methods.
\end{itemize}

\section{Related Work}

\subsection{Machine Unlearning and Association-Level Targets}

Machine unlearning aims to remove specified information from a trained model
without full retraining~\cite{cao2015towards,bourtoule2021machine}. In vision,
standard unlearning targets are usually defined by a forget set at the sample,
class, or concept level. Representative methods include negative-gradient forgetting, SCRUB, SalUn, and
SSD~\cite{golatkar2020eternal,kurmanji2023towards,fan2024salun,foster2024ssd}.

Beyond standard sample-, class-, and concept-level unlearning, recent work has
begun to study unlearning-style interventions for spurious associations and
shortcut dependence~\cite{mitchell2024ule,hakemi2025singleweight,sun2025clear}.
These settings motivate an association-level target: disabling a learned
relation between the task label and a spurious attribute, rather than removing a
sample, class, or attribute in isolation. Our work studies this target from an
evaluation perspective, asking whether such relations are functionally disabled
or merely suppressed.

\subsection{Shortcut Mitigation for Label--Attribute Associations}

Shortcut mitigation methods aim to improve robustness when labels are spuriously
correlated with attributes. Representative methods include GroupDRO, DFR, and JTT, which use group-robust
optimization, classifier retraining, or example reweighting
~\cite{sagawa2020groupdro,kirichenko2023dfr,liu2021jtt}. These methods reduce shortcut-driven behavior
through robustness objectives or reweighting rather than specifying a
relation-level forget target.

Other shortcut-mitigation approaches reduce reliance on spurious attributes
through biased-model failure signals, inferred environments, or contrastive
objectives~\cite{nam2020lff,creager2021environment,zhang2022correct}. These methods can reduce shortcut-driven errors, but output robustness alone
does not determine whether the underlying label--attribute association remains
encoded or functionally restorable. We therefore audit both shortcut-mitigation references and association-adapted
unlearning variants.

\subsection{Evaluating Unlearning Beyond Output Metrics}

Standard unlearning evaluations focus on output behavior: class-unlearning methods are typically
judged by forget-set and retain-set accuracy, while shortcut and association
settings often use subgroup accuracy or worst-group
accuracy~\cite{sagawa2020groupdro,kirichenko2023dfr}. These metrics capture
current prediction behavior, but they do not determine whether the target
information remains encoded in the representation or can be functionally
recovered.

Recent work has therefore examined unlearning beyond output metrics.
Representational audits test whether supposedly removed information remains
readable in hidden features. For example, Gao et al. use linear probes and
nearest-class-center classifiers to show that forgotten-class information can
remain separable after class unlearning~\cite{gao2026illusion}. Yu et al.
evaluate remaining forget-set information through probe recovery, alignment,
separability, and layer-wise analyses~\cite{yu2026mirage}. These works show
that output forgetting does not imply feature-level removal. However, feature
readability alone does not establish that the retained information remains
usable by the model's original classifier head.

Functional restoration methods provide a stronger test by applying lightweight
recovery procedures and measuring whether forgotten behavior returns. Existing
studies recover forgotten behavior through prototype relearning, post-hoc
recovery from residual output signals, or inference-time feature steering
~\cite{ha2025blindspots,qiu2025illusionforgetting,jang2026suppression}. These
approaches reveal that models can satisfy output-level forgetting metrics while
retaining recoverable information. However, they remain largely class-centric:
they primarily test whether a forgotten class or concept can be recovered.

Association unlearning raises a different restoration question: whether a
retained label--attribute relation can be reactivated, rather than whether a
forgotten class can be recovered. \textsc{ART} fills this association-level gap
by auditing functional restorability under the original classifier head, while
probes provide a companion measure of feature readability.

\section{Association Restoration Test (ART)}
\label{sec:method}

\begin{figure*}[t]
    \centering
    \includegraphics[width=\textwidth]{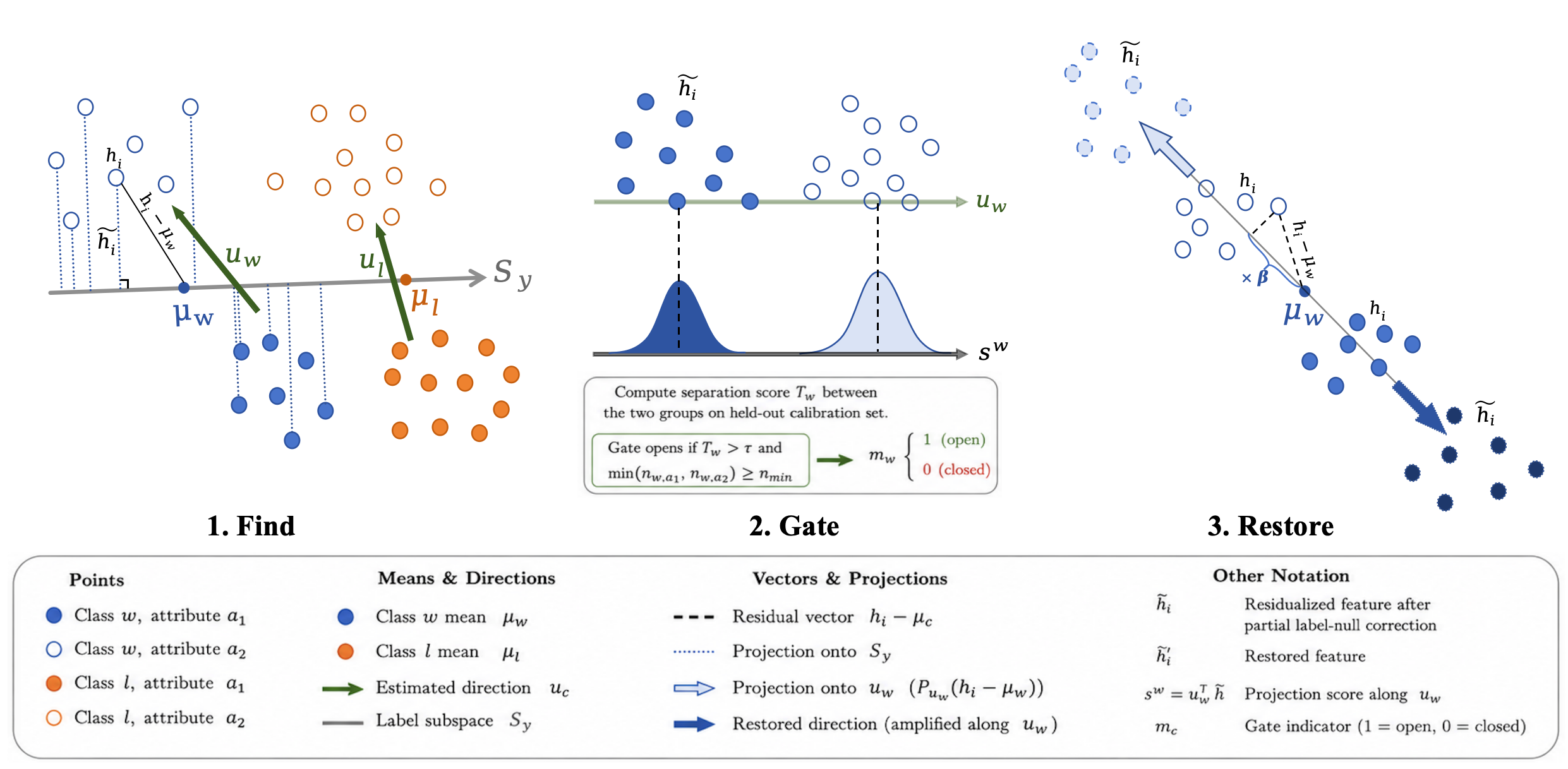}
    \caption{Overview of ART. ART estimates class-conditional association
directions, gates unreliable directions, and amplifies residual shortcut
components before applying the original classifier head.}
    \label{fig:algorithm}
\end{figure*}

\textsc{ART} is a post-hoc feature-space diagnostic for functional
restorability of label--attribute associations. Given a trained classifier, ART
\textbf{finds} class-conditional association directions in frozen feature
space, \textbf{gates} unreliable directions, and \textbf{restores} each
example's residual component along the corresponding direction before applying
the original classifier head. A strong restoration effect indicates that the association remains functionally
connected to the classifier.

\paragraph{Setup.}
In the binary setting, each example is a tuple $(x_i,y_i,a_i)$, where
$y_i\in\mathcal Y=\{y_1,y_2\}$ is the target label and
$a_i\in\mathcal A=\{a_1,a_2\}$ is the spurious attribute. Each subgroup is a
label--attribute pair \((y,a)\). Let \(\pi(a)\) denote the label most associated
with attribute value \(a\) in the biased training distribution. Pairs with
\(y=\pi(a)\) are \emph{aligned}, and pairs with \(y\neq \pi(a)\) are
\emph{conflicting}.

Let \(f=\psi\circ\phi\) be a trained model, where
\(h_i=\phi(x_i)\in\mathbb R^d\) is the frozen representation and \(\psi\) is
the classifier head. ART uses a labeled audit split to estimate association directions and gates, and
a disjoint test split to evaluate restored predictions.

\paragraph{Find.}
For each class \(c\), ART estimates a residual direction that separates the two
attribute groups within that class. Conditioning on \(y=c\) removes the trivial
between-class difference, so the remaining separation reflects within-class
attribute structure.

Let \(\mu_c=\mathbb E[h\mid y=c]\) be the class mean, and let
\(\bar\mu=\mathbb E[h]\) be the global feature mean, both estimated on the
audit split. We define the global label subspace as
\[
    S_y
    =
    \operatorname{span}\{\mu_c-\bar\mu: c\in\mathcal Y\}.
\]
In the binary case, this is equivalent to the one-dimensional span of the
between-class mean difference. Let \(P_{S_y}\) denote the orthogonal projection
onto this subspace. We center features within class \(c\) and apply a partial
label-null correction:
\[
    \tilde h_i^{(c)}
    =
    (I-\rho P_{S_y})(h_i-\mu_c),
    \qquad y_i=c.
\]
Here, \(\rho\in[0,1]\) controls the
strength of label-null correction. We use \(\rho=0.5\) by default to reduce
label-direction leakage while preserving shortcut signal.

ART then contrasts the two attribute groups inside class $c$:
\begin{align*}
    u_c
    &=
    \operatorname{norm}
    (
    \mathbb E[\tilde h^{(c)}\mid y=c,a=a_2]\\
    &-
    \mathbb E[\tilde h^{(c)}\mid y=c,a=a_1]
    ),
\end{align*}
where $\operatorname{norm}(v)=v/(\|v\|_2+\epsilon)$. The direction \(u_c\) captures within-class attribute structure that ART later
tests for functional restorability.

\paragraph{Gate.}
To avoid restoring noisy or poorly supported directions, ART gates each
candidate direction using its projected subgroup separation. On the audit split, we project residual features from
class \(c\) onto \(u_c\):
\[
    s_i^{(c)} = u_c^\top \tilde h_i^{(c)} .
\]
Let \(\mu_{c,j}^s\) and \((\sigma_{c,j}^s)^2\) be the mean and variance of
\(s^{(c)}\) for subgroup \((c,a_j)\). We compute the  separation
score \(T_c\)
between the projected subgroups by:
\[
    T_c
    =
    \frac{
        \left|\mu_{c,2}^{s}-\mu_{c,1}^{s}\right|
    }{
        \sqrt{
        \frac{1}{2}
        \left[
            (\sigma_{c,1}^{s})^2
            +
            (\sigma_{c,2}^{s})^2
        \right]
        + \epsilon}
    } .
\]
Therefore, the gate is
\[
    m_c =
    \mathbf{1}
    \left[
    T_c > \tau
    \;\wedge\;
    \min\left(n_{c,a_1}, n_{c,a_2}\right) \ge n_{\min}
    \right],
\]
where \(n_{c,a_j}\) is the number of audit examples in subgroup \((c,a_j)\). The gate keeps only directions with sufficient projected separation and enough
examples in both attribute groups.
If \(m_c=0\), ART applies no restoration for class \(c\). This prevents noisy or
poorly supported directions from being treated as restorable shortcuts.

\paragraph{Restore.}
Finally, ART tests whether the residual association remains usable by the
original classifier head. For each test example, ART selects the direction using its true label,
\(c_i=y_i\), and amplifies the gated residual component:
\[
    h_i'(\beta)
    =
    h_i+\beta\,m_{c_i}\,P_{U_{c_i}}(h_i-\mu_{c_i}),
    \qquad
    U_c=\operatorname{span}(u_c).
\]
Here, \(\beta\) controls restoration strength, and the projection extracts the
example's residual component along the estimated association direction. ART
therefore amplifies shortcut structure already present in the representation
rather than injecting an arbitrary attribute feature.

The restored prediction is computed using the original classifier head:
\[
    \hat y_i'(\beta)=\psi(h_i'(\beta)).
\]
Therefore, ART asks whether this
controlled feature-space restoration can bring back shortcut behavior.

\section{Experimental Setup}
\paragraph{Datasets.}
We audit methods on three binary spurious-correlation benchmarks: Waterbirds,
CelebA, and SpuCoDogs. We also evaluate a multiclass ISIC 2019 timestamp-artifact extension in
Sec.~\ref{subsec:isic_extension}.
Table~\ref{tab:dataset_semantics} summarizes the label--attribute semantics and
biased training-set group counts.
\begin{table}[!t]
\centering
\scriptsize
\setlength{\tabcolsep}{3.2pt}
\renewcommand{\arraystretch}{1.05}
\caption{
\textbf{Binary benchmark structure.}
We report label--attribute semantics and biased training counts, ordered as
$A_1=(y_1,a_1)$, $C_1=(y_1,a_2)$, $C_2=(y_2,a_1)$, and $A_2=(y_2,a_2)$,
where $A/C$ denote aligned/conflicting groups.
}
\label{tab:dataset_semantics}
\begin{tabular}{@{}llllc@{}}
\toprule
Dataset & Label $y$ & Attribute $a$ & Aligned groups &  Train $(A_1/C_1/C_2/A_2)$ \\
\midrule
Waterbirds
& land / water bird
& land / water bg.
& \makecell[l]{landbird--land\\waterbird--water}
& $3498/184/56/1057$ \\

CelebA
& non-blond / blond
& male / female
& \makecell[l]{non-blond--male\\blond--female}
& $6687/7162/138/2300$ \\

SpuCoDogs
& small / big dog
& indoor / outdoor
& \makecell[l]{small--indoor\\big--outdoor}
& $10000/500/500/10000$ \\
\bottomrule
\end{tabular}
\end{table}

\paragraph{Methods under audit.}
All methods use an ImageNet-pretrained ResNet-50. We include ERM as the biased
reference model and Balanced Retrain as a group-balanced reference trained from
the ImageNet initialization. Because there is no standardized suite of
association-unlearning baselines, we audit methods that target the same
label--attribute shortcuts.

We audit shortcut-mitigation references GroupDRO, DFR, and
JTT~\cite{sagawa2020groupdro,kirichenko2023dfr,liu2021jtt}. These are not
unlearning algorithms, but they reduce shortcut-driven behavior through robust
optimization, classifier retraining, or example reweighting. We also audit association-adapted unlearning variants A-NegGrad+, A-SCRUB,
A-SalUn, and A-SSD~\cite{golatkar2020eternal,kurmanji2023towards,fan2024salun,foster2024ssd}. The prefix ``A-'' denotes an association-specific
adaptation of the original unlearning mechanism. Instead of defining the forget
target as a sample set or class, we define it through a class-conditional
association objective that suppresses predictability of the spurious attribute
from frozen features within each target class, while preserving task-label
performance. A-NegGrad+ replaces the forget-set loss with this association loss;
A-SCRUB keeps the teacher--student preservation term but uses the association
loss as the adversarial forget term; A-SalUn computes saliency with respect to
the association loss before applying masked updates; and A-SSD replaces
forget-set importance with association-direction importance before selective
dampening. Thus, A-* variants preserve the original mechanisms while changing the target
to label--attribute association suppression.

\paragraph{Evaluation metrics.}
\label{sec:metrics}

We use representation probes and \textsc{ART} as complementary diagnostics:
probes test feature readability, while ART tests functional restorability under
the original classifier head.

\textbf{Representational probes.}
We use two standard post-hoc probes on frozen penultimate features: a linear
probe (LP)~\cite{alain2016understanding} and a nearest class-center classifier
(NCC)~\cite{benshaul2022nearest}. LP trains a linear readout, while NCC predicts
by nearest probe-class centroid. Both are trained after model training and leave
the original model unchanged.

We report held-out probe accuracy for three targets: the global spurious
attribute (a), the class-conditional attribute (a) within each target class
(y=c), and the joint subgroup identity ((y,a)). The class-conditional
attribute probe is our main readability diagnostic because association
unlearning does not require globally erasing the attribute; it asks whether
attribute structure remains within each label class.

\textbf{\textsc{ART} restoration metrics.}
To evaluate functional restorability, we compare model behavior before and
after applying \textsc{ART}. Our main output robustness metric is worst-group
accuracy,
\[
    \mathrm{WGA}
    =
    \min_{(y,a)\in\mathcal Y\times\mathcal A}
    \mathrm{Acc}(y,a).
\]
We also measure shortcut-consistent errors on conflicting groups. Let
\[
    \mathcal D_{\mathrm{conf}}
    =
    \{(x_i,y_i,a_i)\in\mathcal D: y_i\neq \pi(a_i)\}.
\]
The conflict shortcut rate is
\[
    \mathrm{CSR}
    =
    \frac{1}{|\mathcal D_{\mathrm{conf}}|}
    \sum_{(x_i,y_i,a_i)\in\mathcal D_{\mathrm{conf}}}
    \mathbf 1[\hat y_i=\pi(a_i)] .
\]
Higher CSR indicates more shortcut-consistent errors on conflicting examples.

We measure the effect of \textsc{ART} using the WGA drop and CSR gain after
restoration:
\[
    \Delta_{\mathrm{WGA}}
    =
    \mathrm{WGA}_{\beta=0}-\mathrm{WGA}_{\beta},
    \qquad
    \Delta_{\mathrm{CSR}}
    =
    \mathrm{CSR}_{\beta}-\mathrm{CSR}_{\beta=0}.
\]
Larger \(\Delta_{\mathrm{WGA}}\) or \(\Delta_{\mathrm{CSR}}\) indicates stronger
functional restoration of shortcut behavior under the original classifier head.

\paragraph{Training and ART protocol.}
Binary benchmark experiments use an ImageNet-pretrained ResNet-50 with a
two-way classifier head; the ISIC extension uses the corresponding multiclass
head. Checkpoints are selected by validation WGA, and results are averaged over five
seeds.

ART is run post hoc on frozen penultimate 2048-dimensional features. Held-out
training features are split into subgroup-stratified direction and gate sets;
the test set is used only once to evaluate restored predictions. We use
\(\rho=0.5\), \(\tau=0.5\), \(n_{\min}=20\), and \(\beta=2\) by default, and
ablate these choices in Sec.~\ref{subsec:sanity_ablation}.

\section{Results}
\label{sec:results}

\subsection{Initial Output Behavior}
\label{subsec:output_before_art}

Before applying \textsc{ART}, we evaluate standard output behavior. Table~\ref{tab:pre_art_output} reports WGA and CSR
on all three binary benchmarks before restoration. Some methods improve WGA or reduce CSR relative to ERM, indicating fewer
shortcut-driven mistakes before restoration. However, these metrics only measure current behavior; they do not show whether
the shortcut is removed or merely hidden but usable.

\begin{table*}[t]
\centering
\scriptsize
\setlength{\tabcolsep}{5.2pt}
\renewcommand{\arraystretch}{1.12}
\caption{
\textbf{Output behavior before \textsc{ART}.}
Each entry reports WGA / CSR over 5 random seeds, with standard deviations
shown as subscripts. Higher WGA and lower CSR indicate weaker shortcut-driven
behavior before restoration.
}
\label{tab:pre_art_output}
\begin{tabular*}{\textwidth}{@{\extracolsep{\fill}}lccc@{}}
\toprule
\textbf{Method}
& \makecell{\textbf{Waterbirds}\\[0.15em]\footnotesize WGA / CSR}
& \makecell{\textbf{CelebA}\\[0.15em]\footnotesize WGA / CSR}
& \makecell{\textbf{SpuCoDogs}\\[0.15em]\footnotesize WGA / CSR} \\
\midrule
ERM
& \metric{11.8}{3.0}{53.2}{4.0}
& \metric{36.7}{3.5}{13.6}{1.8}
& \metric{33.4}{4.0}{65.7}{4.5} \\

Balanced Retrain
& \metric{50.3}{3.9}{17.3}{4.1}
& \metric{80.6}{3.5}{15.3}{3.9}
& \metric{77.4}{4.0}{13.2}{2.1} \\

\midrule
DFR
& \metric{81.3}{2.5}{17.2}{2.0}
& \metric{81.7}{1.8}{15.3}{1.5}
& \metric{86.0}{1.5}{13.2}{1.2} \\

GroupDRO
& \metric{64.0}{4.0}{26.0}{3.0}
& \metric{82.6}{1.8}{16.6}{1.5}
& \metric{85.6}{1.8}{12.3}{1.3} \\

JTT
& \metric{30.3}{2.9}{47.8}{3.2}
& \metric{78.5}{3.4}{10.8}{3.7}
& \metric{83.0}{3.9}{31.7}{2.4} \\

\midrule
A-NegGrad+
& \metric{38.9}{5.0}{49.6}{4.5}
& \metric{81.8}{2.0}{17.4}{1.8}
& \metric{70.6}{3.0}{26.8}{2.8} \\

A-SCRUB
& \metric{28.3}{5.0}{48.2}{4.5}
& \metric{57.2}{5.5}{13.7}{2.0}
& \metric{55.8}{5.0}{44.2}{4.0} \\

A-SalUn
& \metric{20.6}{4.5}{34.3}{4.0}
& \metric{82.8}{1.8}{16.7}{1.5}
& \metric{68.0}{3.5}{29.6}{3.0} \\

A-SSD
& \metric{36.8}{4.5}{32.8}{3.5}
& \metric{81.1}{2.0}{18.3}{1.8}
& \metric{64.6}{4.0}{31.8}{3.5} \\

\bottomrule
\end{tabular*}
\begin{flushleft}
\scriptsize
All values are percentages. WGA denotes worst-group accuracy, and CSR denotes
the conflict shortcut rate.
\end{flushleft}
\end{table*}

\subsection{Feature Readability under Post-hoc Probes}
\label{subsec:representational_probes}

We probe frozen penultimate features using LP and NCC readouts. Table~\ref{tab:probe_summary} reports LP / NCC accuracies for three probe
targets: the class-conditional attribute, the global attribute, and the joint
group. 

Across datasets, probes show little evidence of representational deletion. Class-conditional attribute scores remain close to ERM, with small standard
deviations across methods. This indicates that the spurious attribute remains readable within each target
class. Thus, the worst group accuracy improvements of the methods in
Sec.~\ref{subsec:output_before_art} generally do not correspond to shortcut removal from feature space. However, this result
alone does not imply functional failure: a readable shortcut may still be
decoupled from the original classifier head.

\begin{table}[tb]
\centering
\scriptsize
\setlength{\tabcolsep}{3pt}
\renewcommand{\arraystretch}{1.0}
\caption{
\textbf{Representational probes on frozen penultimate features.}
Each entry reports LP / NCC accuracy. The audited average is computed over the
non-retrain audited methods and excludes ERM and Balanced Retrain.
}
\label{tab:probe_summary}
\begin{tabular}{@{}llcc@{}}
\toprule
\textbf{Dataset} & \textbf{Probe} & \textbf{ERM} & \textbf{Audited avg.} \\
\midrule
\multirow{3}{*}{Waterbirds}
& Cond-Attr   & $85.0 / 89.0$ & \probe{84.6}{0.7}{89.4}{0.7} \\
& Global-Attr & $89.3 / 88.4$ & \probe{88.7}{0.4}{88.0}{0.8} \\
& Group       & $64.4 / 76.0$ & \probe{64.7}{0.5}{76.4}{0.7} \\
\midrule
\multirow{3}{*}{CelebA}
& Cond-Attr   & $77.6 / 78.8$ & \probe{77.4}{0.8}{78.0}{1.8} \\
& Global-Attr & $85.9 / 78.3$ & \probe{85.8}{0.7}{80.2}{1.4} \\
& Group       & $63.8 / 70.8$ & \probe{63.8}{0.8}{70.2}{1.1} \\
\midrule
\multirow{3}{*}{SpuCoDogs}
& Cond-Attr   & $95.1 / 94.1$ & \probe{95.2}{0.5}{93.2}{1.5} \\
& Global-Attr & $97.1 / 91.5$ & \probe{97.1}{0.2}{87.8}{4.2} \\
& Group       & $78.9 / 74.9$ & \probe{79.7}{1.3}{74.2}{1.6} \\
\bottomrule
\end{tabular}

\begin{flushleft}
\scriptsize
\emph{Abbreviations:} LP = linear probe; NCC = nearest class-center classifier;
Cond-Attr = class-conditional attribute probe. All values are percentages.
\end{flushleft}
\end{table}

\subsection{Functional Restoration with ART}
\label{subsec:main_art_results}

\begin{table*}[t]
\centering
\small
\setlength{\tabcolsep}{7pt}
\renewcommand{\arraystretch}{1.12}
\caption{
\textbf{\textsc{ART} restoration effects.}
Each cell reports \(\Delta_{\mathrm{WGA}} / \Delta_{\mathrm{CSR}}\) at
\(\beta=2\), where
\(\Delta_{\mathrm{WGA}}=\mathrm{WGA}_{\beta=0}-\mathrm{WGA}_{\beta=2}\) and
\(\Delta_{\mathrm{CSR}}=\mathrm{CSR}_{\beta=2}-\mathrm{CSR}_{\beta=0}\).
Larger values indicate stronger functional restoration of
shortcut-consistent behavior under the original classifier head. Bold
\(\Delta_{\mathrm{WGA}}\) values indicate drops of at least 20 points.
}
\label{tab:main_art_restoration_summary}

\begin{tabular}{@{}lccc@{}}
\toprule
\textbf{Method}
& \makecell{\textbf{Waterbirds}\\ \(\Delta_{\mathrm{WGA}} / \Delta_{\mathrm{CSR}}\)}
& \makecell{\textbf{CelebA}\\ \(\Delta_{\mathrm{WGA}} / \Delta_{\mathrm{CSR}}\)}
& \makecell{\textbf{SpuCoDogs}\\ \(\Delta_{\mathrm{WGA}} / \Delta_{\mathrm{CSR}}\)} \\
\midrule

ERM
& \restpair{7.3}{2.5}{35.9}{3.5}
& \restpairbold{22.8}{3.2}{17.2}{3.2}
& \restpairbold{31.4}{1.8}{32.0}{1.5} \\

Balanced Retrain
& \restpair{3.0}{4.5}{-7.0}{2.0}
& \restpair{2.7}{2.2}{9.8}{2.5}
& \restpair{2.2}{2.0}{-10.0}{1.2} \\

\midrule
GroupDRO
& \restpair{8.7}{4.5}{17.6}{4.0}
& \restpairbold{23.7}{4.0}{12.3}{3.0}
& \restpair{10.8}{2.5}{-4.3}{2.0} \\

DFR
& \restpair{9.2}{3.0}{38.2}{4.5}
& \restpairbold{24.5}{4.0}{15.6}{3.5}
& \restpair{11.4}{2.2}{15.6}{3.5} \\

JTT
& \restpair{7.1}{3.8}{18.0}{4.2}
& \restpairbold{26.8}{6.0}{18.4}{3.7}
& \restpair{11.4}{3.9}{11.3}{2.4} \\

\midrule
A-NegGrad+
& \restpairbold{28.2}{4.0}{37.0}{4.5}
& \restpairbold{35.7}{5.0}{19.0}{4.2}
& \restpairbold{36.2}{5.5}{34.4}{5.0} \\

A-SCRUB
& \restpair{19.9}{3.5}{38.4}{4.5}
& \restpairbold{32.8}{5.5}{19.8}{4.5}
& \restpairbold{45.8}{4.5}{44.4}{4.5} \\

A-SalUn
& \restpair{12.8}{3.0}{39.4}{4.2}
& \restpairbold{28.9}{4.5}{17.7}{3.8}
& \restpairbold{44.4}{5.0}{40.3}{5.0} \\

A-SSD
& \restpairbold{20.1}{4.5}{42.6}{4.5}
& \restpairbold{47.8}{6.0}{21.4}{4.5}
& \restpairbold{21.0}{4.8}{20.5}{5.0} \\

\bottomrule
\end{tabular}
\begin{flushleft}
\scriptsize
Each entry is reported in percentage points as
\((\Delta_{\mathrm{WGA}} \pm \mathrm{std}) /
(\Delta_{\mathrm{CSR}} \pm \mathrm{std})\). Positive
\(\Delta_{\mathrm{WGA}}\) means WGA decreases after \textsc{ART}; positive
\(\Delta_{\mathrm{CSR}}\) means shortcut-consistent errors increase after
\textsc{ART}.
\end{flushleft}
\end{table*}

We apply ART to test whether residual shortcut associations are functionally
restorable. Table~\ref{tab:main_art_restoration_summary} reports WGA drop and
CSR gain at the default setting \((\beta=2,\rho=0.5)\).

Table~\ref{tab:main_art_restoration_summary} shows that ART separates
low-restoration references from methods with functionally restorable shortcut
associations. Balanced Retrain provides the strongest low-restoration reference, with small
WGA drops and little or negative CSR gain. DFR and GroupDRO are often less vulnerable than the association-adapted unlearning variants, although they are not uniformly stable across all
datasets. In contrast, A-NegGrad+, A-SCRUB, A-SalUn, and A-SSD frequently show
large WGA drops together with large CSR gains, especially on Waterbirds and
SpuCoDogs.  Thus, many methods improve output behavior while
leaving residual associations that ART can reactivate. 

\begin{figure*}[t]
\centering

\begin{minipage}[t]{0.4\textwidth}
    \centering
    \includegraphics[width=\linewidth]{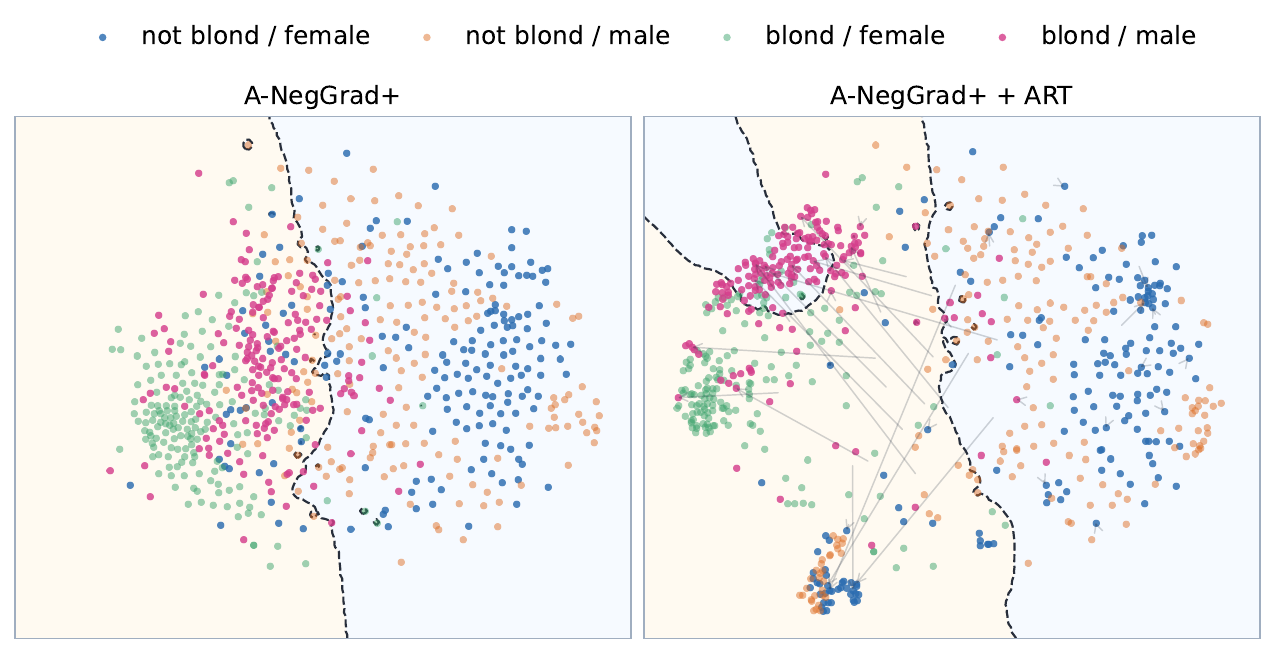}\\
    {\scriptsize (a) A-NegGrad+ feature space}

    \includegraphics[width=\linewidth]{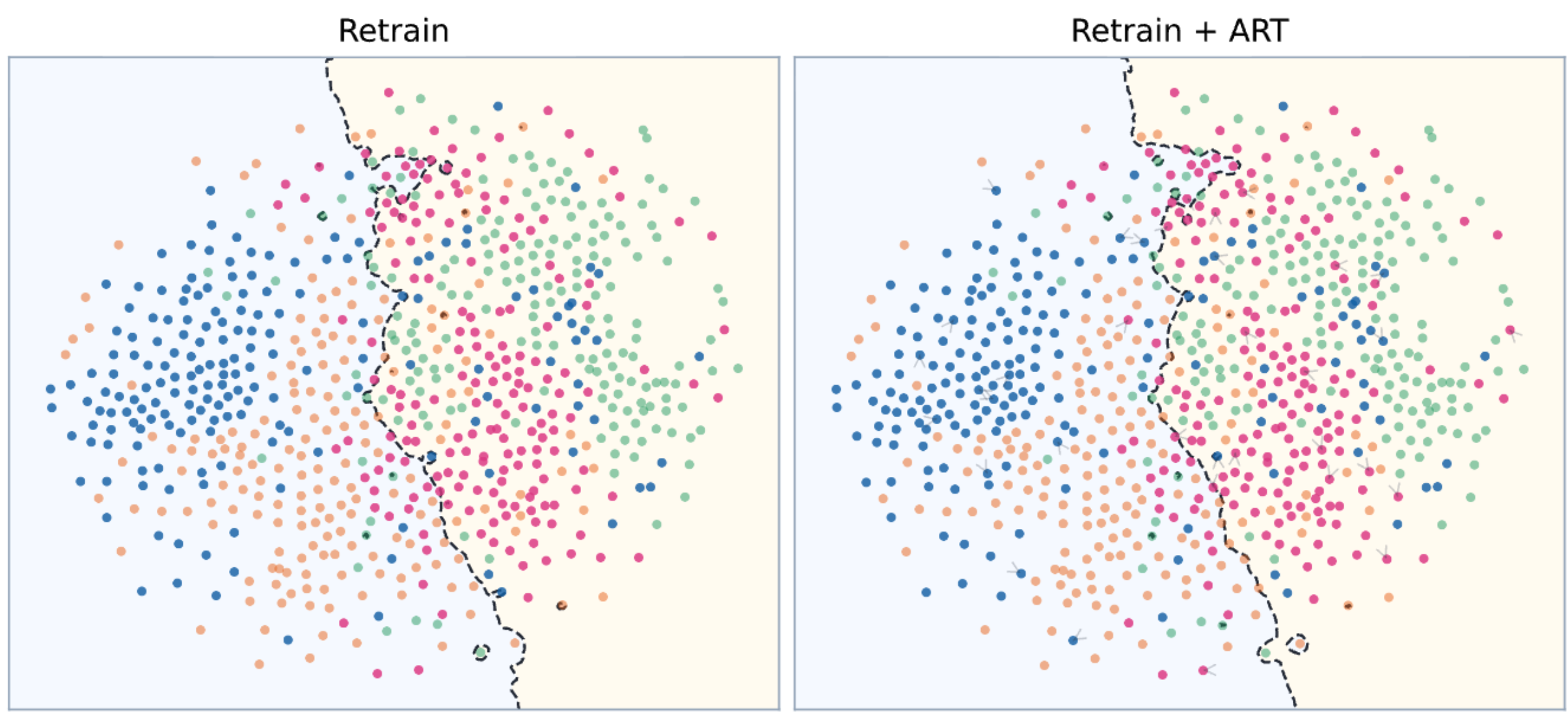}\\
    {\scriptsize (b) Retrain feature space}
\end{minipage}
\hfill
\begin{minipage}[t]{0.5\textwidth}
    \centering
    \includegraphics[width=\linewidth]{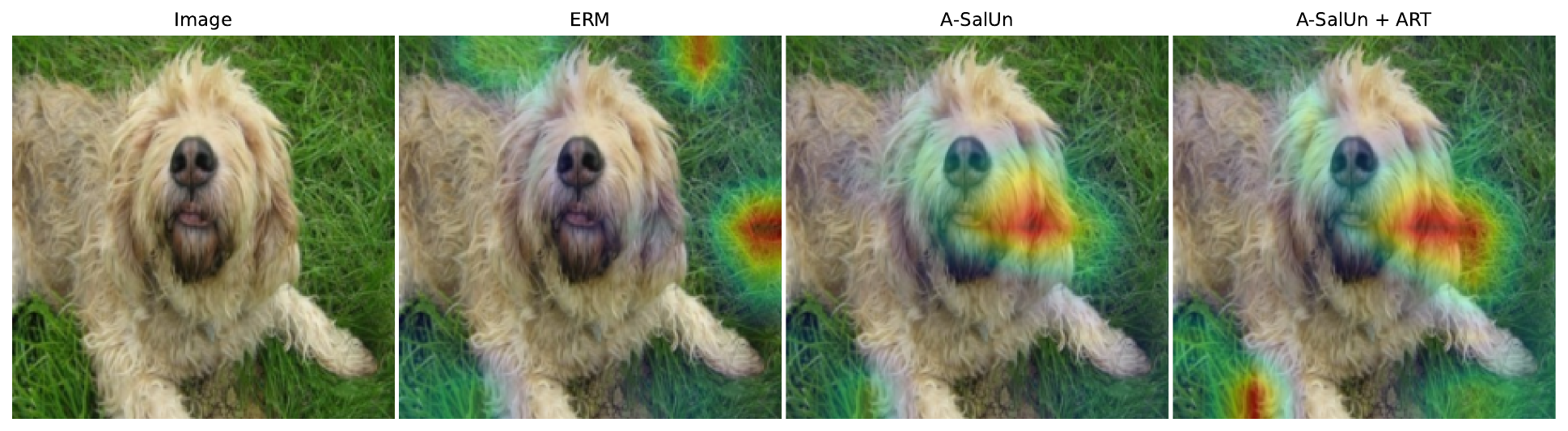}\\
    {\scriptsize (c) A-SalUn Grad-CAM}

    \includegraphics[width=\linewidth]{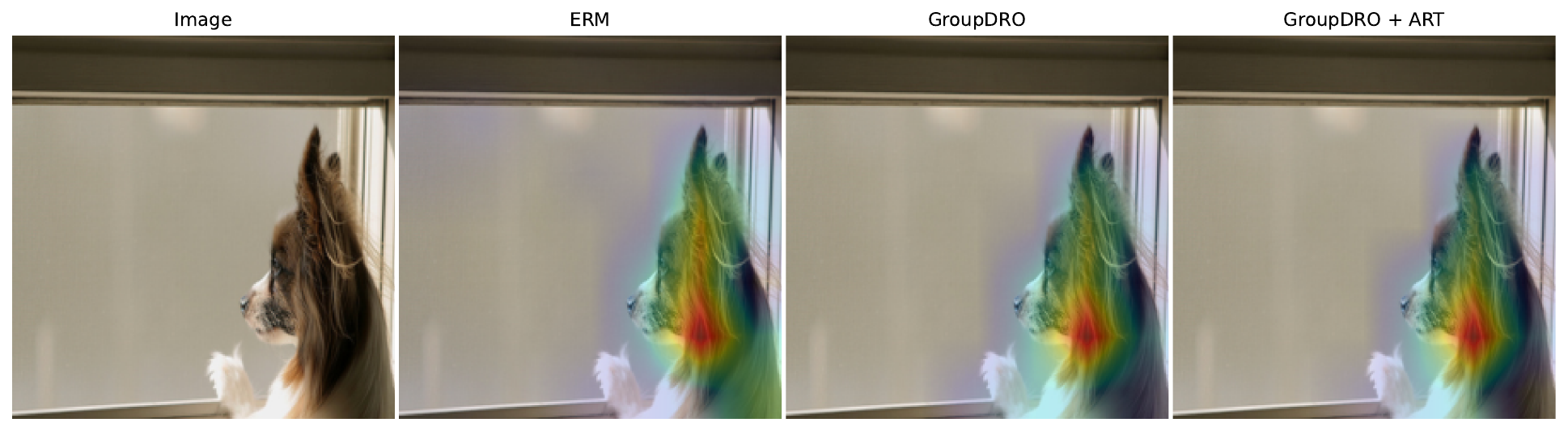}\\
    {\scriptsize (d) GroupDRO Grad-CAM}
\end{minipage}

\caption{
\textbf{Qualitative visualizations under ART.}
Left: t-SNE visualizations of CelebA penultimate features before and after ART
restoration. Right: Grad-CAM visualizations on SpuCoDogs. 
}
\label{fig:qualitative_art_visualizations}
\end{figure*}

\paragraph{Qualitative illustration.}
Fig.~\ref{fig:qualitative_art_visualizations} shows representative stable and
vulnerable cases. In feature space, A-NegGrad+
exhibits a larger ART-induced shift than Retrain. In Grad-CAM visualizations,
ART increases saliency on shortcut-associated background regions for A-SalUn,
whereas GroupDRO remains more focused on the object. These visualizations illustrate the quantitative WGA and CSR patterns.

\subsection{Joint Interpretation of Readability and Restorability}
\label{subsec:final_taxonomy}

\begin{figure}[tb]
\centering
\includegraphics[width=0.55\linewidth]{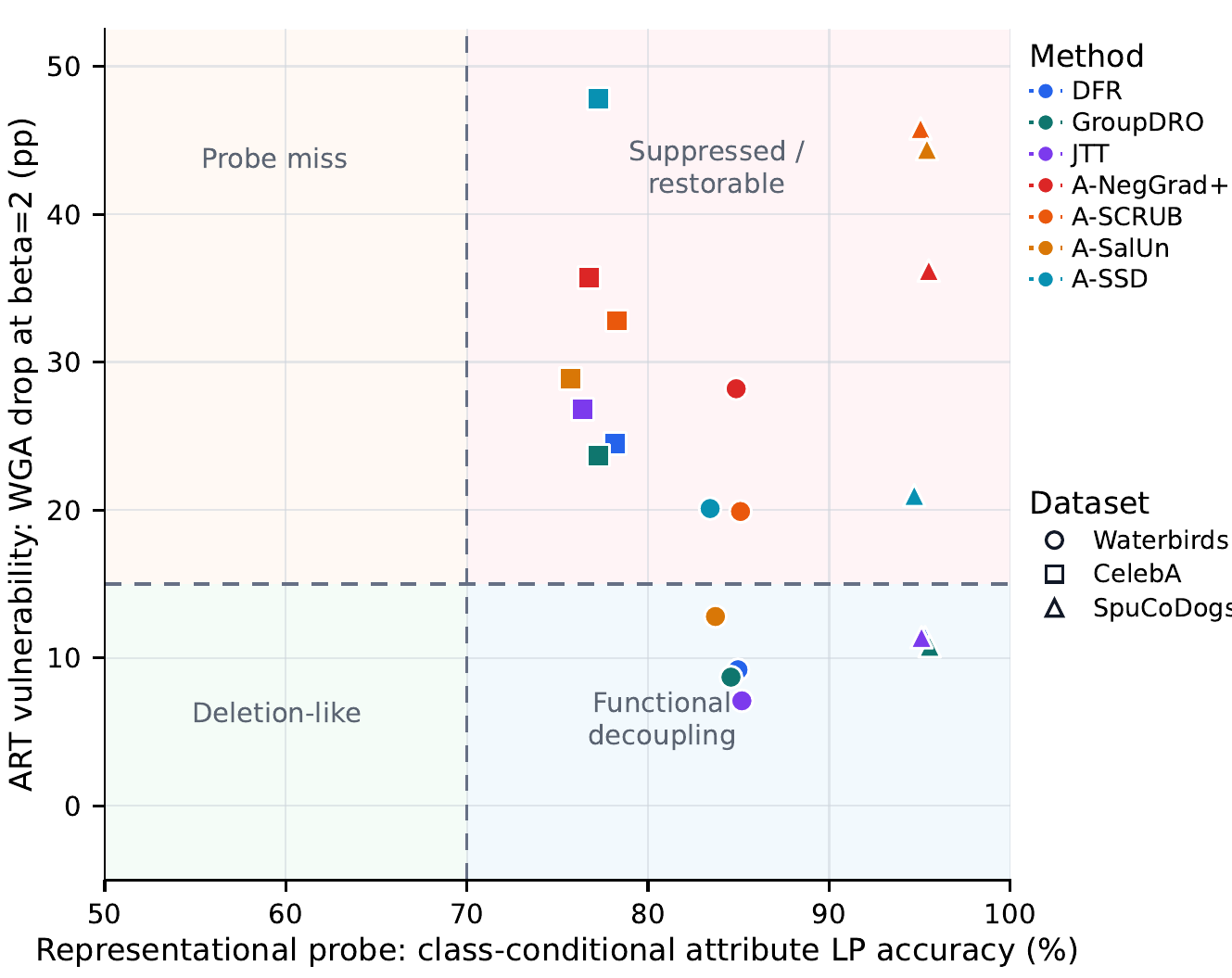}
\caption{
Probe--ART taxonomy. Each point is a method--dataset pair, excluding ERM and Retrain.
}
\label{fig:probe_art_taxonomy_scatter}
\end{figure}

Fig.~\ref{fig:probe_art_taxonomy_scatter} compares feature readability and ART
restorability. The x-axis is class-conditional attribute probe accuracy; the
y-axis is WGA drop at \(\beta=2\). We use this plot as a qualitative taxonomy rather
than a hard decision rule. We treat probe accuracy above \(70\%\) as clear feature-space readability and a WGA drop above \(15\) points as a substantial restoration effect. Together, these guides divide the plot into four regions. Low probe accuracy and low ART vulnerability suggest \textit{representational deletion}: the association is neither readable from the representation nor easily restored into model behavior. High probe accuracy but low ART vulnerability suggests \textit{functional decoupling}: the association is still encoded in the features, but the classifier head cannot easily restore it. High probe accuracy and high ART vulnerability indicate \textit{restorable shortcuts}: the association remains readable and can still be reactivated into shortcut-consistent predictions. Finally, low probe accuracy but high ART vulnerability indicates a possible probe miss, where the association is not captured by the probe but can still affect the classifier through the ART intervention.

Most methods fall in the high-probe region, indicating little evidence of
representational deletion. DFR and GroupDRO are closer to functional
decoupling in several settings, whereas A-NegGrad+, A-SCRUB, A-SalUn, and
A-SSD often fall in the suppressed/restorable region. Thus, similar output improvements can reflect different internal states: some
methods reduce the functional influence of retained shortcuts, while others
preserve associations that ART can reactivate.

\subsection{Sanity Checks and Ablations}
\label{subsec:sanity_ablation}

\paragraph{Control perturbations.}
Control perturbations test whether ART reflects association-specific
restoration rather than generic feature-space disruption. We compare ART with controls that use the same operation but replace the
association direction with shuffled or random directions.
Table~\ref{tab:art_control_perturbations} shows that these controls produce
near-zero WGA drops, while ART produces much larger drops across datasets. Thus, the effect depends on the estimated class-conditional association
direction, not generic feature perturbation.

\begin{table}[t]
\centering
\scriptsize
\setlength{\tabcolsep}{4pt}
\renewcommand{\arraystretch}{1.08}
\caption{
\textbf{Control perturbations for ART.}
Each entry reports mean WGA drop with standard deviation / maximum drop at
\(\beta=2\). Negative-control directions produce near-zero drops, indicating
that ART is not a generic feature-space perturbation.
}
\label{tab:art_control_perturbations}
\begin{tabular}{@{}lccc@{}}
\toprule
\textbf{Intervention}
& \textbf{Waterbirds}
& \textbf{CelebA}
& \textbf{SpuCoDogs} \\
\midrule
ART
& \dropstat{16.5}{7.6}{28.2}
& \dropstat{32.2}{8.9}{47.8}
& \dropstat{28.3}{16.0}{45.8} \\

Shuffled-\(a\) forced
& \dropstat{0.6}{0.1}{2.2}
& \dropstat{1.2}{0.9}{2.8}
& \dropstat{0.3}{0.2}{3.2} \\

Random direction
& \dropstat{0.0}{0.1}{0.3}
& \dropstat{0.1}{0.0}{0.6}
& \dropstat{0.1}{0.1}{0.6} \\

Random matched subspace
& \dropstat{-0.1}{0.2}{0.5}
& \dropstat{0.2}{0.1}{1.1}
& \dropstat{-0.1}{0.3}{0.4} \\

\bottomrule
\end{tabular}
\end{table}

\paragraph{Hyperparameters.}
Our default ART configuration uses \(\beta=2\) and \(\rho=0.5\). We use \(\beta=2\) as a restoration stress test because it makes stability
differences clearly measurable. We set \(\rho=0.5\) as a partial label-null correction to reduce leakage from
the global label direction without making the association estimate overly
conservative.

We ablate both choices to test sensitivity to this default setting. Increasing \(\beta\) consistently strengthens ART restoration;
for example, on CelebA, the mean WGA drop increases from \(4.2\) to \(32.2\)
points as \(\beta\) increases from \(0.25\) to \(2\). In contrast, increasing
\(\rho\) weakens restoration by removing more label-aligned variation before
direction estimation; for example, on CelebA, the mean WGA drop decreases from
\(38.7\) points at \(\rho=0\) to \(1.9\) points at \(\rho=1\). Thus,
\(\rho=0.5\) provides a useful tradeoff between suppressing label leakage and
preserving shortcut signal.

\paragraph{Design choices.}
We ablate three design choices in the ART find--gate--restore pipeline, as
summarized in Table~\ref{tab:component_ablation}. All entries are averaged over non-retrain audited methods.

For \emph{Find}, we compare the proposed class-conditional association
direction with a global attribute direction and a label direction. The
class-conditional ART direction gives the clearest shortcut-restoration signal:
it produces large WGA drops while also yielding the largest CSR gains across all
three datasets. In contrast, the label direction can sometimes reduce WGA, but
it produces little or no CSR increase, indicating that it does not specifically
restore shortcut-consistent errors. 

For \emph{Gate}, we compare the default gated version with an ungated variant.
Removing the gate makes the intervention more aggressive and can increase the
WGA drop, but it does not consistently increase shortcut-consistent restoration.
We therefore use the gate as a conservative specificity control that filters
weak or poorly supported directions.

For \emph{Restore}, we compare true-label selection with predicted-label
selection. Predicted-label ART remains effective, but it is consistently weaker
because misclassified examples can be routed to the wrong class-conditional
direction. Thus, true-label selection provides a cleaner controlled diagnostic,
while the predicted-label variant shows that the restoration effect does not
depend entirely on oracle label routing.

\begin{table*}[t]
\centering
\scriptsize
\setlength{\tabcolsep}{4.5pt}
\renewcommand{\arraystretch}{1.12}
\caption{
\textbf{Component ablations for ART.}
We ablate the three stages of the ART pipeline: \textit{Find}, \textit{Gate},
and \textit{Restore}. Each cell is formatted as mean WGA drop\(_{\pm \mathrm{std}}\) /
mean CSR gain\(_{\pm \mathrm{std}}\). All values are percentage points.
Averages exclude ERM and Balanced Retrain.
}
\label{tab:component_ablation}
\begin{tabular}{@{}llccc@{}}
\toprule
\textbf{Step}
& \textbf{Variant}
& \textbf{Waterbirds}
& \textbf{CelebA}
& \textbf{SpuCoDogs} \\
\midrule

\multirow{3}{*}{Find}
& Class-conditional ART
& \dirres{16.5}{7.6}{35.5}{9.0}
& \dirres{32.2}{8.9}{17.6}{3.3}
& \dirres{28.3}{16.0}{25.2}{18.3} \\

& Global-Attr direction
& \dirres{11.5}{7.0}{20.0}{11.0}
& \dirres{20.5}{7.5}{8.5}{4.0}
& \dirres{15.0}{10.5}{11.5}{12.0} \\

& Label direction
& \dirres{18.0}{8.5}{2.5}{5.0}
& \dirres{24.0}{8.0}{1.5}{4.0}
& \dirres{28.0}{12.0}{-1.0}{6.0} \\

\midrule

\multirow{2}{*}{Gate}
& With gate
& \gatepred{16.5}{7.6}{35.5}{9.0}
& \gatepred{32.2}{8.9}{17.6}{3.3}
& \gatepred{28.3}{16.0}{25.2}{18.3} \\

& Without gate
& \gatepred{21.0}{13.0}{29.0}{18.0}
& \gatepred{36.0}{11.0}{16.0}{6.5}
& \gatepred{30.0}{19.0}{20.0}{23.0} \\

\midrule

\multirow{2}{*}{Restore}
& True-label selection
& \labelsel{16.5}{7.6}{35.5}{9.0}
& \labelsel{32.2}{8.9}{17.6}{3.3}
& \labelsel{28.3}{16.0}{25.2}{18.3} \\

& Predicted-label selection
& \labelsel{13.2}{8.4}{23.0}{13.0}
& \labelsel{25.5}{9.0}{12.0}{5.5}
& \labelsel{18.5}{14.0}{14.5}{17.0} \\

\bottomrule
\end{tabular}
\end{table*}

\subsection{Extension to Multiclass Associations}
\label{subsec:isic_extension}

We also evaluate ART on a multiclass ISIC2019 skin-lesion classification
setting~\cite{tschandl2018ham10000,codella2018skin,hernandez2024bcn20000}.
We create a controlled shortcut by adding date stamps more often to MEL
training images, encouraging a timestamp--MEL association. At evaluation, each test image is run in clean and timestamped form. We compute WGA over groups defined by class and
timestamp presence, and measure the timestamp shortcut rate by checking how
often timestamped non-MEL images are incorrectly predicted as MEL.

ART extends naturally to this setting by estimating a separate timestamp
direction within each lesion class. The results mirror the binary benchmarks. Balanced Retrain remains largely stable, suggesting the timestamp association
is not easily restored after balanced training. In contrast, several audited methods still show
restoration effects. DFR and A-SSD show the clearest WGA degradation, with drops
of \(31.8\) and \(17.9\) points, respectively. Other methods, such as GroupDRO
and A-NegGrad+, show restoration mainly through increased timestamp-to-MEL
shortcut errors rather than WGA degradation. A-SCRUB and A-SalUn show smaller
but still non-negligible increases in this shortcut rate. Overall, the ISIC
extension suggests that ART can reveal restorable shortcut associations beyond
binary label--attribute settings.

\subsection{ART-Targeted Mitigation}
\label{subsec:mitigation}

Finally, we test whether ART directions can serve as a head-level robustness
signal. Starting from A-NegGrad+, we freeze the backbone and retrain only the
final linear head using both clean and ART-restored features. This head-only
procedure does not remove representation-level shortcuts; instead, it tests
whether ART can guide the classifier away from functionally restorable
associations.

Table~\ref{tab:art_head_mitigation} shows that the base A-NegGrad+ head is
highly vulnerable: applying ART causes large WGA drops and substantial CSR
increases. After retraining the classifier head on both clean and ART-restored
features, this vulnerability is largely removed. In Waterbirds and CelebA, the
ART-perturbed features even yield higher WGA than clean features, producing
negative \(\Delta\). We interpret this not as representation-level shortcut
deletion, but as head-level functional decoupling: the retrained head no longer
maps the ART-amplified association direction to shortcut-consistent predictions.
This benefit may come at a cost, as seen in CelebA where clean WGA decreases.
Thus, ART-head is a mitigation probe demonstrating that ART directions can guide
classifier decoupling, rather than a complete association-unlearning method.

\begin{table}[t]
\centering
\scriptsize
\setlength{\tabcolsep}{3.2pt}
\renewcommand{\arraystretch}{1.08}
\caption{
\textbf{ART-targeted head mitigation.}
Clean and ART denote WGA before and after ART at \(\beta=2\). 
\(\Delta\) is Clean\(-\)ART. Lower \(\Delta\) and smaller CSR increase indicate
reduced functional restorability.
}
\label{tab:art_head_mitigation}
\begin{tabular}{@{}llrrrr@{}}
\toprule
\textbf{Dataset}
& \textbf{Model}
& \textbf{Clean}
& \textbf{ART}
& \textbf{\(\Delta\)}
& \textbf{CSR \(0 \rightarrow 2\)} \\
\midrule
\multirow{2}{*}{Waterbirds}
& A-NegGrad+ base
& 38.9 & 10.7 & 28.2 & 49.6 \(\rightarrow\) 86.5 \\
& ART-head
& 52.8 & 64.6 & -11.8 & 28.5 \(\rightarrow\) 23.3 \\
\midrule
\multirow{2}{*}{CelebA}
& A-NegGrad+ base
& 81.8 & 46.1 & 35.7 & 17.4 \(\rightarrow\) 36.4 \\
& ART-head
& 50.6 & 71.1 & -20.6 & 12.3 \(\rightarrow\) 6.1 \\
\midrule
\multirow{2}{*}{SpuCoDogs}
& A-NegGrad+ base
& 70.6 & 34.4 & 36.2 & 26.8 \(\rightarrow\) 61.2 \\
& ART-head
& 79.2 & 74.8 & 4.4 & 20.6 \(\rightarrow\) 24.8 \\
\bottomrule
\end{tabular}
\end{table}

\section{Conclusion, Limitations, and Future Work}
\label{sec:conclusion}

We introduced ART as a post-hoc diagnostic for testing whether retained
label--attribute associations remain usable by the original classifier head. Across datasets, class-conditional shortcut structure often remains readable in
frozen features even when output robustness improves.
ART shows that readable associations differ in restorability: some methods
decouple the head from retained shortcut structure, while others leave
reactivatable associations that produce WGA drops and CSR gains.

Our work has several limitations. Low ART vulnerability indicates only that the specific restoration tested by ART does not readily reactivate the association; alternative nonlinear interventions may still recover it. ART also requires an audit split with target and attribute labels, and our
implementation uses linear class-conditional directions in penultimate feature
space.
Other layers, architectures, nonlinear structures, or real-world multi-attribute
associations may require different restoration estimators.

Future work should develop methods that more directly eliminate shortcut
associations from internal representations, rather than primarily decoupling
them from the classifier head. ART should also be extended beyond linear penultimate-space directions to
nonlinear, layer-dependent, and multi-attribute shortcuts. Extending restoration-aware association audits to generative and
vision-language models is another promising direction.

%
%
\clearpage
\bibliographystyle{splncs04}
\bibliography{main}
\end{document}